\documentclass[runningheads]{llncs}
\usepackage{graphicx}
\usepackage{amsmath,amssymb} % define this before the line numbering.
\usepackage{color}
\usepackage{multirow}
\usepackage{enumitem}
\usepackage{caption}
\usepackage{amsmath,bm}

\usepackage{authblk}
\usepackage[symbol*]{footmisc}

\DefineFNsymbolsTM{otherfnsymbols}{%
	\textdagger    \dagger
	\textdaggerdbl \ddagger
	\textbardbl    \|%
	\textparagraph \mathparagraph
}

\setfnsymbol{otherfnsymbols}

\usepackage{makecell}
\usepackage{float}
\usepackage[dvipsnames]{xcolor}

\usepackage[width=122mm,left=12mm,paperwidth=146mm,height=193mm,top=12mm,paperheight=217mm]{geometry}

\usepackage{array}
\newcolumntype{L}[1]{>{\raggedright\let\newline\\\arraybackslash\hspace{0pt}}m{#1}}
\newcolumntype{C}[1]{>{\centering\let\newline\\\arraybackslash\hspace{0pt}}m{#1}}
\newcolumntype{R}[1]{>{\raggedleft\let\newline\\\arraybackslash\hspace{0pt}}m{#1}}

\begin{document}
% \linenumbers
\pagestyle{headings}
\mainmatter

\title{GeoDesc: Learning Local Descriptors by Integrating Geometry Constraints} 
% Replace with your title

\titlerunning{GeoDesc: Learning Local Descriptors by Integrating Geometry Constraints}
% Replace with a meaningful short version of your title

\authorrunning{Z. Luo, T. Shen, L. Zhou, S. Zhu, R. Zhang, Y. Yao, T. Fang and L. Quan}
% Replace with shorter version of the author list. If there are more authors than fits a line, please use A. Author et al.

\iffalse
\author{
	Zixin Luo$^{1}$\thanks{\footnotesize Zixin Luo, Lei Zhou and Yao Yao were summer interns at Everest Innovation Technology (Altizure).}\orcidID{0000-0001-6946-2826}, Tianwei Shen$^1$\thanks{\footnotesize Tianwei Shen and Runze Zhang were interns at Everest Innovation Technology (Altizure).}\orcidID{0000-0002-3290-2258}, ~~~~~~~~~
	Lei Zhou$^{1\dagger}$\orcidID{0000-0003-4988-5084}, Siyu Zhu$^{1}$\thanks{\footnotesize Siyu Zhu is with Alibaba A.I. Labs since Oct. 2017.}\orcidID{0000-0003-0293-0044},  ~~~~~~~~~~~~~
	Runze Zhang$^{1\ddagger}$\thanks{\footnotesize Runze Zhang is the corresponding author.}\orcidID{0000-0001-9698-0178}, Yao Yao$^{1\dagger}$\orcidID{0000-0001-9866-4291}, ~~~~~~~ Tian Fang$^{2}$\orcidID{0000-0000-0000-0000}, Long Quan$^{1}$\orcidID{0000-0001-8148-1771}
 }
\fi

\author{
	Zixin Luo$^{1}$, Tianwei Shen$^1$,
	Lei Zhou$^{1}$, Siyu Zhu$^{1}$\thanks{\footnotesize Siyu Zhu is with Alibaba A.I. Labs since Oct. 2017.},
	\\ Runze Zhang$^{1}$, Yao Yao$^{1}$, Tian Fang$^{2}$, Long Quan$^{1}$
}

%Please write out author names in full in the paper, i.e. full given and family names. 
%If any authors have names that can be parsed into FirstName LastName in multiple ways, please include the correct parsing, in a comment to the volume editors:
%\index{Lastnames, Firstnames}
%(Do not uncomment it, because you may introduce extra index items if you do that, we will use scripts for introducing index entries...)

\institute{
	$^1$~Hong Kong University of Science and Technology, \\
	\email{ \{zluoag,tshenaa,lzhouai,szhu,rzhangaj,yyaoag,quan\}@cse.ust.hk} \\
	$^2$~Shenzhen Zhuke Innovation Technology (Altizure), \\
	\email{fangtian@altizure.com}
}

\maketitle

\begin{abstract}
	Learned local descriptors based on Convolutional Neural Networks (CNNs) have achieved significant improvements on patch-based benchmarks, whereas not having demonstrated strong generalization ability on recent benchmarks of image-based 3D reconstruction. In this paper, we mitigate this limitation by proposing a novel local descriptor learning approach that integrates geometry constraints from multi-view reconstructions, which benefits the learning process in terms of data generation, data sampling and loss computation. We refer to the proposed descriptor as GeoDesc, and demonstrate its superior performance on various large-scale benchmarks, and in particular show its great success on challenging reconstruction tasks. Moreover, we provide guidelines towards practical integration of learned descriptors in Structure-from-Motion (SfM) pipelines, showing the good trade-off that GeoDesc delivers to 3D reconstruction tasks between accuracy and efficiency.
	

\end{abstract}

\section{Introduction}
Computing local descriptors on interest regions serves as the subroutine of various computer vision applications such as panorama stitching~\cite{li2015dual}, wide baseline matching~\cite{Mishkin2015WxBS}, image retrieval~\cite{Philbin2007Object}, and Structure-from-Motion (SfM)~\cite{Zhu_2018_CVPR,shen2016graph,zhang2017distributed,zhu2014local}. A powerful descriptor is expected to be invariant to both photometric and geometric changes, such as illumination, blur, rotation, scale and perspective changes. Due to the reliability, efficiency and portability, hand-crafted descriptors such as SIFT~\cite{Lowe:2004kp} have been influentially dominating this field for more than a decade. Until recently, great efforts have been made on developing learned descriptors based on Convolutional Neural Networks (CNNs), which have achieved surprising results on patch-based benchmarks such as HPatches dataset~\cite{Balntas:2017vx}. However, on image-based datasets such as ETH local features benchmark~\cite{schonberger2017comparative}, learned descriptors are found to underperform advanced variants of hand-crafted descriptors. The contradictory findings raise the concern of integrating those purportedly better descriptors in real applications, and show significant room of improvement for developing more powerful descriptors that generalize to a wider range of scenarios.

One possible cause of above contradictions, as demonstrated in~\cite{schonberger2017comparative}, is the lack of generalization ability as a consequence of data insufficiency. Although previous research~\cite{Balntas:un,SimoSerra:2015waa,Sohn:2016wj} discusses several effective sampling methods that produce seemingly large amount of training data, the generalization ability is still bounded to limited data sources, e.g., the widely-used Brown dataset~\cite{Brown:2011dd} with only 3 image sets. Hence, it is not surprising that resulting descriptors tend to overfit to particular scenarios. To overcome it, research such as~\cite{Tian:2017uv,zhang2017learning} applies extra regularization for compact feature learning. Meanwhile, LIFT~\cite{Yi:2016te} and \cite{mitra2018large} seek to enhance data diversity and generate training data from reconstructions of Internet tourism data. However, the existing limitation has not yet been fully mitigated, while intermediate geometric information is overlooked in the learning process despite the robust geometric property that local patch preserves, e.g., the well approximation of local deformations~\cite{morel2009asift}. 

Besides, we lack guidelines for integrating learned descriptors in practical pipelines such as SfM. In particular, the \emph{ratio criterion}, as suggested in~\cite{Lowe:2004kp} and justified in~\cite{kaplan2016interpreting}, has received almost no individual attention or was considered inapplicable for learned descriptors~\cite{schonberger2017comparative}, whereas it delivers excellent matching efficiency and accuracy improvements, and serves as the necessity for pipelines such as SfM to reject false matches and seed feasible initialization. A general method to apply ratio criterion for learned descriptors is in need in practice.  

In this paper, we tackle above issues by presenting a novel learning framework that takes advantage of geometry constraints from multi-view reconstructed data. In particular, we address the importance of data sampling for descriptor learning and summarize our contributions threefold. i) We propose a novel batch construction method that simulates the pair-wise matching and effectively samples useful data for learning process. ii) Collaboratively, we propose a loss formulation to reduce overfitting and improve the performance with geometry constraints. iii) We provide guidelines about ratio criterion, compactness and scalability towards practical portability of learned descriptors. 

We evaluate the proposed descriptor, referred to as GeoDesc, on traditional \cite{Heinly:2012dd} and recent two large-scale datasets~\cite{Balntas:2017vx,schonberger2017comparative}. Superior performance is shown over the state-of-the-art hand-crafted and learned descriptors. We mitigate previous limitations by showing consistent improvements on both patch-based and image-based datasets, and further demonstrate its success on challenging 3D reconstructions.
\section{Related Works}

\smallskip\noindent\textbf{Networks design.} Due to weak semantics and efficiency requirements, existing descriptor learning often relies on shallow and thin networks, e.g., three-layer networks in DDesc~\cite{SimoSerra:2015waa} with 128-dimensional output features. Moreover, although widely-used in high-level computer vision tasks, max pooling is found to be unsuitable for descriptor learning, which is then replaced by L2 pooling in DDesc~\cite{SimoSerra:2015waa} or even removed in L2-Net~\cite{Tian:2017uv}. To further incorporate scale information, DeepCompare~\cite{zagoruyko2015learning} and L2-Net~\cite{Tian:2017uv} use a two-stream central-surround structure which delivers consistent improvements at extra computational cost. To improve the rotational invariance, an orientation estimator is proposed in~\cite{Yi:2015ty}. Besides of feature learning, previous efforts are also made on joint metric learning as in~\cite{zagoruyko2015learning,Han:2015gv,G:2016dj}, whereas comparison in Euclidean space is more preferable by recent works~\cite{Balntas:un,SimoSerra:2015waa,Tian:2017uv,Yi:2016te,Balntas:2016vs} in order to guarantee its efficiency.

\smallskip\noindent\textbf{Loss formulation}
Various of loss formulations have been explored for effective descriptor learning. Initially, networks with a learned metric use softmax loss~\cite{zagoruyko2015learning,Han:2015gv} and cast the descriptor learning to a binary classification problem (similar/dissimilar). With weakly annotated data, \cite{markuvs2016learning} formulates the loss on keypoint bags. More generally, pair-wise loss~\cite{SimoSerra:2015waa,Yi:2016te} and triplet loss~\cite{Balntas:un,G:2016dj,Balntas:2016vs} are used by networks without a learned metric. Both loss formulations encourage matching patches to be close whereas non-matching patches to be far-away in some measure space. In particular, triplet loss delivers better results~\cite{Balntas:un,G:2016dj} as it suffers less overfitting~\cite{lin2015deephash}. For effective training, recent L2-Net~\cite{Tian:2017uv} and HardNet~\cite{Mishchuk:2017tj} use the structured loss for data sampling which drastically improves the performance. To further boost the performance, extra regularizations are introduced for learning compact representation in~\cite{Tian:2017uv,zhang2017learning}.

\smallskip\noindent\textbf{Evaluation protocol}
Previous works often evaluate on datasets such as~\cite{Mikolajczyk:2005tk,Winder2009Picking,Heinly:2012dd}. However, those datasets either are small, or lack diversity to generalize well to various applications of descriptors. As a result, the evaluation results are commonly inconsistent or even contradictory to each other as pointed out in~\cite{Balntas:2017vx}, which limits the application of learned descriptors. Two novel benchmarks, HPatches~\cite{Balntas:2017vx} and ETH local descriptor benchmark~\cite{schonberger2017comparative} have been recently introduced with clearly defined protocols and better generalization properties. However, inconsistency still exists in the two benchmarks, where HPatches~\cite{Balntas:2017vx} benchmark demonstrates the significant outperformance from learned descriptors over the handcrafted, whereas the ETH local descriptor benchmark~\cite{schonberger2017comparative} finds that the advanced variants of the traditional descriptor are at least on par with the learning-based. The inconclusive results indicate that there is still significant room for improvement to learn more powerful feature descriptors.

\section{Method}

\subsection{Network architecture}

We borrow the network in L2-Net~\cite{Tian:2017uv}, where the feature tower is constructed by eschewing pooling layers and using strided convolutional layers for in-network downsampling. Each convolutional layer except the last one is followed by a batch normalization (BN) layer whose weighting and bias parameters are fixed to $1$ and $0$. The L2-normalization layer after the last convolution produces the final $128$-dimensional feature vector.
% The configuration is shown in Table~\ref{tab:net_arch}.
\iffalse
\begin{table}[h]
	\centering
	\caption{Parameters of the feature tower: a $32\times 32$ grayscale patch input yields a 128-dimensional normalized feature vector}
	\label{tab:net_arch}
	\resizebox{0.58\textwidth}{!}{
		\begin{tabular}{ccccc}
			\Xhline{1pt}
			& \textbf{Input size} & \textbf{Filter size} & \textbf{Channels} & \textbf{Stride} \\ \Xhline{0.7pt}
			\emph{Conv. + BN} & $32\times 32$ & $3\times 3$ & 32 & 1 \\
			\emph{Conv. + BN} & $32\times 32$ & $3\times 3$ & 32 & 1 \\
			\emph{Conv. + BN} & $32\times 32$ & $3\times 3$ & 64 & 2 \\
			\emph{Conv. + BN} & $16\times 16$ & $3\times 3$ & 64 & 1 \\
			\emph{Conv. + BN} & $16\times 16$ & $3\times 3$ & 128 & 2 \\
			\emph{Conv. + BN} & $8\times 8$ & $3\times 3$ & 128 & 1 \\
			\emph{Conv.} & $8\times 8$ & $8\times 8$ & 128 & 1 \\
			\emph{L2-Norm.} & $128$ & - & - & - \\
			\Xhline{1pt}
		\end{tabular}
	}
\end{table}
\fi

\subsection{Training data generation}

\label{sec:data_praperation}
Acquiring high quality training data is important in learning tasks. In this section, we discuss a practical pipeline that automatically produces well-annotated data suitable for descriptor learning.

\smallskip\noindent\textbf{2D correspondence generation.} Similar to LIFT~\cite{Yi:2016te}, we rely on successful 3D reconstructions to generate ground truth 2D correspondences in an automatic manner. First, sparse reconstructions are obtained from standard SfM pipeline~\cite{Sch2016Structure}. Then, 2D correspondences are generated by projecting 3D point clouds. In general, SfM is used to filter out most mismatches among images.

\begin{figure}[h]
	\centering 
	\includegraphics[width=\textwidth]{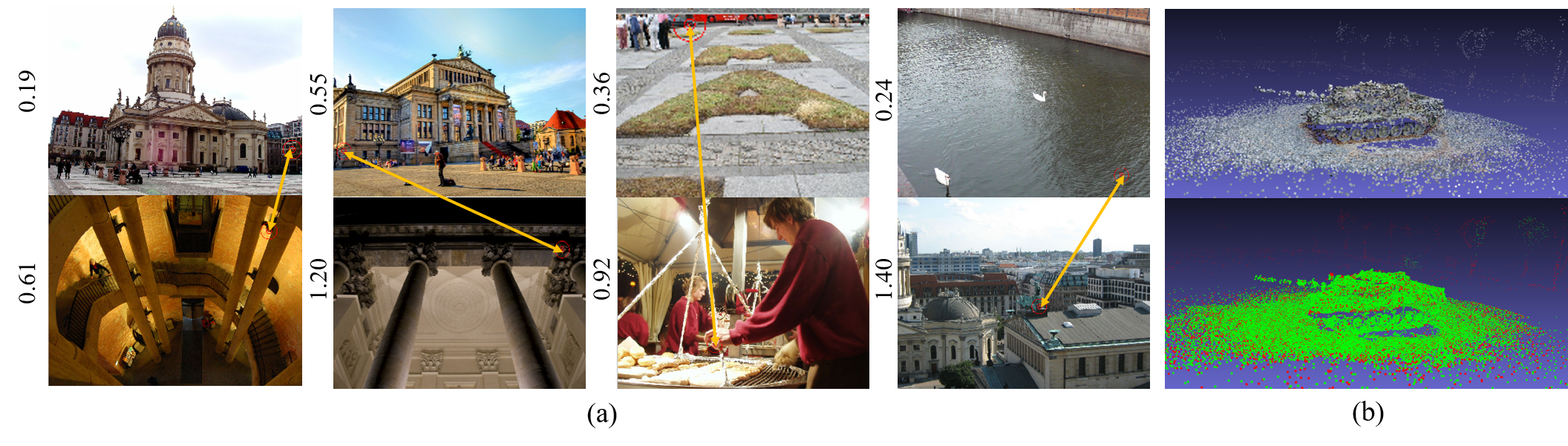}
	\caption{(a) Outlier matches after SfM verification (by COLMAP~\cite{Sch2016Structure}) on \emph{Gendarmenmarkt} dataset~\cite{Wilson:2014wq}. The reprojection error (next to the image) cannot be used to identify false matches. (b) Reconstructed sparse point cloud (top), where points in red (bottom) indicate being filtered by Delaunay triangulation and only reliable points in green are kept. The number of points decreases from 75k to 53k after the filtering.}
	\label{fig:outlier}
\end{figure}

Although verified by SfM, the generated correspondences are still outlier-contaminated from image noise and wrongly registered cameras. It happens particularly often on Internet tourism datasets such as~\cite{Wilson:2014wq,Radenovic:2016ks} (illustrated in Fig.~\ref{fig:outlier}(a)), and usually not likely to be filtered by simply limiting reprojection error. To improve data quality, we take one step further than LIFT by computing the visibility check based on 3D Delaunay triangulation~\cite{labatut2007efficient} which is widely-used for outlier filtering in dense stereo. Empirically, $30\%$ of 3D points will be discarded after the filtering while only points with high precision are kept for ground truth generation. Fig.~\ref{fig:outlier}(b) gives an example to illustrate its effect.

\smallskip\noindent\textbf{Matching patch generation.} Next, the interest region of a 2D projection is cropped similar to LIFT, which is formulated by an similarity transformation
\begin{equation}\label{eq:sampling}
\begin{matrix}
\begin{pmatrix}
x_i^s \\ y_i^s
\end{pmatrix}
=
\begin{bmatrix}
\frac{k\sigma}{2}cos(\theta) & \frac{k\sigma}{2}sin(\theta) & x \\ -\frac{k\sigma}{2}sin(\theta) & \frac{k\sigma'}{2}cos(\theta) & y
\end{bmatrix}
\begin{pmatrix}
x_i^t \\ y_i^t
\end{pmatrix}
\end{matrix},
\end{equation}
where $(x^s_i , y_i^s ), (x^t_i , y_i^t)$ are input and output regular sampling grids, and $(x, y, \sigma, \theta)$ are keypoint parameters ($x, y$ coordinates, scale and orientation) from SIFT detector. The constant $k$ is set to $12$ as in LIFT, resulting in $12\sigma\times 12\sigma$ patches.

Due to the robust estimation of scale ($\sigma$) and orientation ($\theta$) parameters of SIFT even in extreme cases~\cite{zhou2017progressive}, the resulting patches are mostly free of scale and rotation differences, thus suitable for training. In later experiments of image matching or SfM, we rely on the same cropping method to achieve scale and rotation invariance for learned descriptors.

\subsection{Geometric similarity estimation} 
\label{sec:geo_sim}
Geometries at a 3D point are robust and provide rich information. Inspired by the MVS (Multi-View Stereo) accuracy measurement in~\cite{Zhang:2015bm}, we define two types of geometric similarity: patch similarity and image similarity, which will facilitate later data sampling and loss formulation. 

\begin{figure}[h]
	\centering 
	\includegraphics[width=\textwidth]{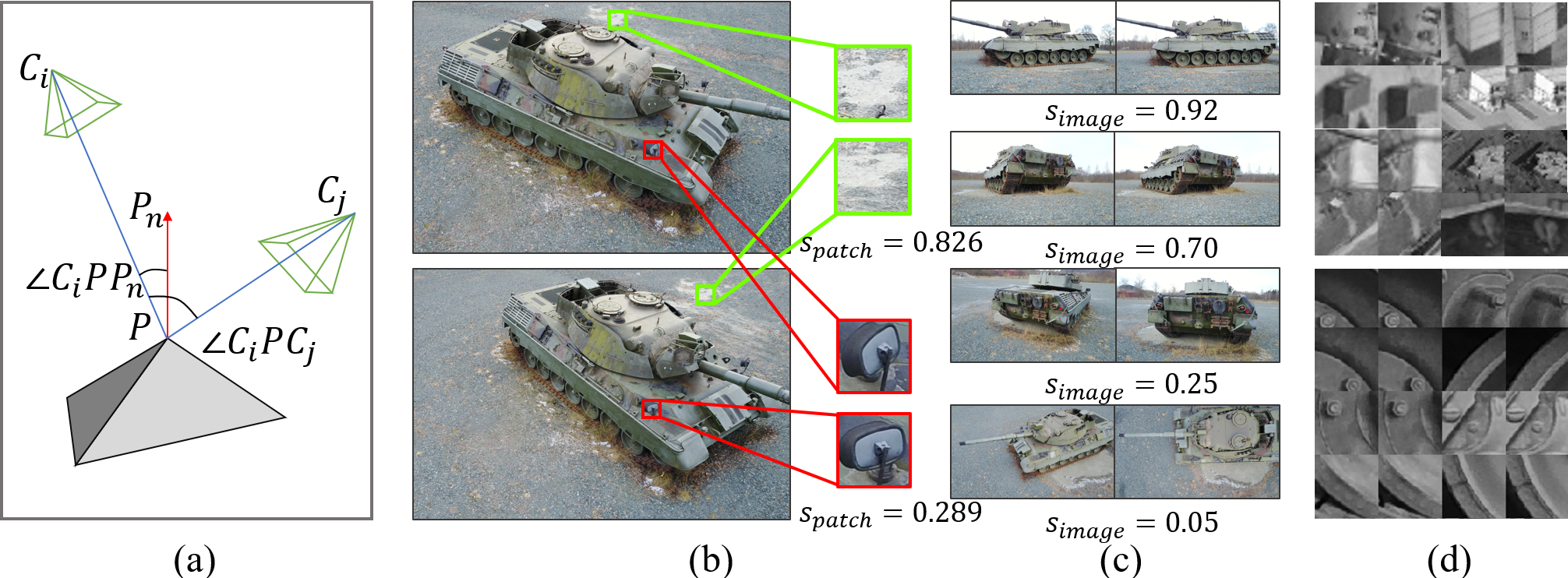}
	\caption{(a) The patch similarity relies on the geometric relationship between cameras, tracks and surface normal. (b) The effect of patch similarity, which measures the difficulty to have a patch pair matched with respect to the perspective change. (c) The effect of image similarity, which measures the perspective change between an image pairs. (d) Batch data constructed by L2-Net~\cite{Tian:2017uv} and HardNet~\cite{Mishchuk:2017tj} (top), whose in-batch patch pairs are often distinctive to each other and thus contribute nothing to the loss in the late learning (e.g., the margin-based loss). However, the batch data from the proposed batch construction method (bottom) consists of  similar patch pairs due to the spatially close keypoints or repetitive patterns, which are considered harder to distinguish and thus raise greater challenges for learning}
	\label{fig:geo_sim}
\end{figure}

\smallskip\noindent\textbf{Patch similarity.} We define patch similarity $S_{patch}$ to measure the difficulty to have a patch pair matched with respect to perspective changes. Formally, given a patch pair, we relate it to its corresponding 3D track $P$ which is seen by  cameras centering at $C_i$ and $C_j$. Next, we compute the vertex normal $P_n$ at $P$ from the surface model. The geometric relationship is illustrated in Fig.~\ref{fig:geo_sim}(a). Finally, we formulate $S_{patch}$ as 
\begin{equation}
S_{patch} = s_1 s_2 = g(\angle C_iPC_j, \sigma_1)g(\angle C_iPP_n - \angle C_jPP_n, \sigma_2),
\end{equation}
where $s_1$ measures the intersection angle between two viewing rays from the 3D track ($\angle C_iPC_j$), while $s_2$ measures the difference of incidence angles between a viewing ray and the vertex normal from the 3D track ($\angle C_iPP_n, \angle C_jPP_n$). The angle metric is defined as $g(\alpha, \sigma) = \exp(-\frac{\alpha^2}{2\sigma^2})$. As an interpretation, $s_1$ and $s_2$ measure the perspective change regarding a \emph{3D point} and local \emph{3D surface}, respectively. The effect of $S_{patch}$ is illustrated in Fig.~\ref{fig:geo_sim}(b). 

The accuracy of $s_1$ and $s_2$ depends on sparse and mesh reconstructions, respectively, and is generally sufficient for its use as shown in~\cite{Zhang:2015bm}. The similarity does not consider scale and rotation changes as already resolved from Equation~\ref{eq:sampling}. Empirically, we choose $\sigma_1 = 15$ and $\sigma_2 = 20$ (in degree).

\smallskip\noindent\textbf{Image similarity.} Based on the patch similarity, we define the image similarity $S_{image}$ as the average patch similarity of the correspondences between an image pair. The image similarity measures the difficulty to match an image pair and can be interpreted as a measurement of perspective change. Examples are given in Fig.~\ref{fig:geo_sim}(c). The image similarity will be beneficial for data sampling in Sec.~\ref{sec:batch_construction}.

\subsection{Batch construction}
\label{sec:batch_construction}
For descriptor learning, most existing frameworks take patch pairs (matching/non-matching) or patch triplets (reference, matching and non-matching) as input. As in previous studies, the convergence rate is highly dependent on being able to see useful data~\cite{MovshovitzAttias:2017wg}. Here, ``useful" data often refers to patch pairs/triplets that produce meaningful loss for learning. However, the effective sampling of such data is generally challenging due to the intractably large number of patch pair/triplet combination in the database. Hence, on one hand, sampling strategies such as hard negative mining~\cite{SimoSerra:2015waa} and anchor swap~\cite{Balntas:un} are proposed, while on the other hand, effective batch construction is used in~\cite{Tian:2017uv,G:2016dj,Mishchuk:2017tj} to compare the reference patch against all the in-batch samples in the loss computation. 

Inspired by previous works, we propose a novel batch construction method that effectively samples ``useful" data by relying on geometry constraints from SfM, including the image matching results and image similarity $S_{image}$, to simulate the pair-wise image matching and sample data. Formally, given one image pair, we extract a \emph{match set} $X=\{(x_1, x^+_1), (x_2, x^+_2), ..., (x_{N_1}, x_{N_1}^+)\}$, where $N_1$ is the set size and $(x_i, x^+_i)$ is a matching patch pair surviving the SfM verification. A training batch is then constructed on $N_2$ match sets. Hence, the learning objective becomes to improve the matching quality for each match set. In Sec.~\ref{sec:loss}, we will discuss the loss computation on each match set and batch data.

% Most existing learned descriptors are trained with batch samples constructed from individual patch pairs. Instead, \cite{markuvs2016learning} optimized the descriptor with weakly-labelled data between image pairs. However, its performance is limited compared with recent L2-Net or HardNet since true and false matches are not correctly identified in its training. In this section, we propose a batch construction method which simulates pair-wise image matching and cooperates with well-labelled patch pairs.

Compared with L2-Net~\cite{Tian:2017uv} and HardNet~\cite{Mishchuk:2017tj} whose training batches are  random sampled from the whole database, the proposed method produces harder samples and thus raises greater challenges for learning. As an example shown in Fig.~\ref{fig:geo_sim}(d), the training batch constructed by the proposed method consists of many similar patterns, due to the spatially close keypoints or repetitive textures. In general, such training batch has two major advantages for descriptor learning:

\begin{itemize}[leftmargin=*, noitemsep, topsep=0pt]
	\item It reflects the in-practice complexity. In real applications, image patches are often extracted between image pairs for matching. The proposed method simulates this scenario so that training and testing become more consistent. 
	
	\item It generates hard samples for training. As observed in~\cite{Balntas:un,SimoSerra:2015waa,Mishchuk:2017tj,MovshovitzAttias:2017wg}, hard samples are critical to fast convergence and performance improvement for descriptor learning. The proposed method effectively generates batch data that is sufficiently hard, while not being overfitting as constructed on real matching results instead of model inference results~\cite{SimoSerra:2015waa}.
\end{itemize}

To further boost the training efficiency, we exclude image pairs that are too similar to contribute to the learning. Those pairs are effectively identified by the image similarity defined in Sec.~\ref{sec:geo_sim}. In practice, we discard image pairs whose $S_{image}$ are larger than $0.85$ (e.g., the toppest pair in Fig.~\ref{fig:geo_sim}(c)), which results in a $30\%$ shrink of training samples. 

\subsection{Loss formulation}
\label{sec:loss}
We formulate the loss with two terms: structured loss and geometric loss. 

\smallskip\noindent\textbf{Structured loss.}
The structured loss used in L2-Net~\cite{Tian:2017uv} and HardNet~\cite{Mishchuk:2017tj} is essentially suitable to consume the batch samples constructed in Sec.~\ref{sec:batch_construction}. In particular, the formulation in HardNet based on the ``hardest-in-batch" strategy and a distance margin shows to be more effective than the log-likelihood formulation in L2-Net. However, we observe successive overfitting when applying the HardNet loss to our batch data, which we ascribe to the too strong constraint of ``hardest-in-batch" strategy. In this strategy, the loss is computed on the data sample that produces the largest loss, and a margin with a large value ($1.0$ in HardNet) is set to push the non-matching pairs away from matching pairs. In our batch data,  we already effectively sample the ``hard" data which is often visually similar, thus forcing a large margin is inapplicable and stalls the learning. One simple solution is to decrease the margin value, whereas the performance drops significantly in our experiments.

To avoid above limitation and better take advantage of our batch data, we propose the loss formulation as follows. First, we compute the structured loss for one match set. Given normalized features $\mathbf{F}_1, \mathbf{F}_2 \in \mathbb{R}^{N_1\times 128}$ computed on match set $X$ for all $(x_i, x_i^+)$, the cosine similarity matrix is derived by $\mathbf{S} = \mathbf{F}_1 \mathbf{F}_2^T$. Next, we compute $\mathbf{L} = \mathbf{S} - \alpha \mathbf{diag}(\mathbf{S})$ and formulate the loss as
\begin{equation}
E_1 = \frac{1}{N_1(N_1-1)}\sum_{i, j}(\max(0, l_{i, j} - l_{i, i}) + \max(0, l_{i, j} - l_{j, j})),
\end{equation}
where $l_{i, j}$ is the element in $\mathbf{L}$, and $\alpha\in(0,1)$ is the distance ratio mimicking the behavior of ratio test~\cite{Lowe:2004kp} and pushing away non-matching pairs from matching pairs. Finally, we take the average of the loss on each match set to derive the final loss for one training batch.

The proposed formulation is distinctive from HardNet in three aspects. First, we compute the cosine similarity instead of Euclidean distance for computational efficiency. Second, we apply a \emph{distance ratio margin} instead of a \emph{fixed distance margin} as an adaptive margin to reduce overfitting. Finally, we compute the \emph{mean} value of each loss element instead of the maximum (``hardest-in-batch") in order to cooperate the proposed batch construction.

\smallskip\noindent\textbf{Geometric loss.} Although $E_1$ ensures matching patch pairs to be distant from the non-matching, it does not explicitly encourage matching pairs to be close in its measure space. One simple solution is to apply a typical pair-wise loss in~\cite{SimoSerra:2015waa}, whereas taking a risk of positive collapse and overfitting as observed in~\cite{lin2015deephash}. To overcome it, we adaptively set up the margin regarding the patch similarity defined in Sec.~\ref{sec:geo_sim}, serving as a soft constraint for maximizing  the positive similarity. We refer to this term as \emph{geometric loss} and formulate it as
\begin{equation}
E_2 = \sum_{i}{\max(0, \beta - s_{i, i})}, \beta = \begin{cases}
0.7 & s_{patch} \geq 0.5 \\
0.5 & 0.2 \leq s_{patch} < 0.5 \\
0.2 & \text{otherwise}
\end{cases}
\end{equation}
where $\beta$ is the adaptive margin, $s_{i, i}$ is the element in $S$, namely, the cosine similarity of patch pair $(x_i, x_i^+)$, while $s_{patch}$ is the patch similarity for $(x_i, x_i^+)$. We use $E_1 + \lambda E_2$ as the final loss, and empirically set $\alpha$ and $\lambda$ to $0.4$ and $0.2$.

\subsection{Training}
We use image sets~\cite{Wilson:2014wq} as in LIFT~\cite{Yi:2016te}, the SfM data in~\cite{Radenovic:2016ks}, and further collect several image sets to form the training database. Based on COLMAP~\cite{Sch2016Structure}, we run 3D reconstructions to establish necessary geometry constraints. Image sets that are overlapping with the benchmark data are manually excluded.
We train the networks from scratch using Adam with a base learning rate of $0.001$ and weight decay of $0.0001$. The learning rate decays by $0.9$ every $10,000$ steps. Data augmentation includes randomly flipping, $90$ degrees rotation and brightness and contrast adjustment. The match set size $N_1$ and batch size $N_2$ are 64 and 12, respectively. Input patches are standardized to have zero mean and unit norm. 

\section{Experiments}
We evaluate the proposed descriptor on three datasets: the patch-based HPatches~\cite{Balntas:2017vx} benchmark, the image-based Heinly benchmark~\cite{Heinly:2012dd} and ETH local features benchmark~\cite{schonberger2017comparative}. We further demonstrate  on challenging SfM examples.

\subsection{HPatches benchmark}
\label{sec:hbench}
HPatches benchmark~\cite{Balntas:2017vx} defines three complementary tasks: patch verification, patch matching, and patch retrieval. Different levels of geometrical perturbations are imposed to form EASY, HARD and TOUGH patch groups. In the task of verification, two subtasks are defined based on whether negative pairs are sampled from images within the same (SAMESEQ) or different sequences (DIFFSEQ). In the task of matching, two subtasks are defined based on whether the principle variance is viewpoint (VIEW) or illumination (ILLUM). Following~\cite{Balntas:2017vx}, we use mean average precision (mAP) to measure the performance for all three tasks on HPatches split `full'.

We select five descriptors to compare: SIFT as the baseline, RSIFT~\cite{Arandjelovic:2012wm} and DDesc~\cite{SimoSerra:2015waa} as the best-performing hand-crafted and learned descriptors concluded in~\cite{Balntas:2017vx}. Moreoever, we experiment with recent learned descriptors L2-Net~\cite{Tian:2017uv} and HardNet~\cite{Mishchuk:2017tj}. The proposed descriptor is referred to as GeoDesc.
 
\begin{figure}[h]
	\centering 
	\includegraphics[width=0.95\textwidth]{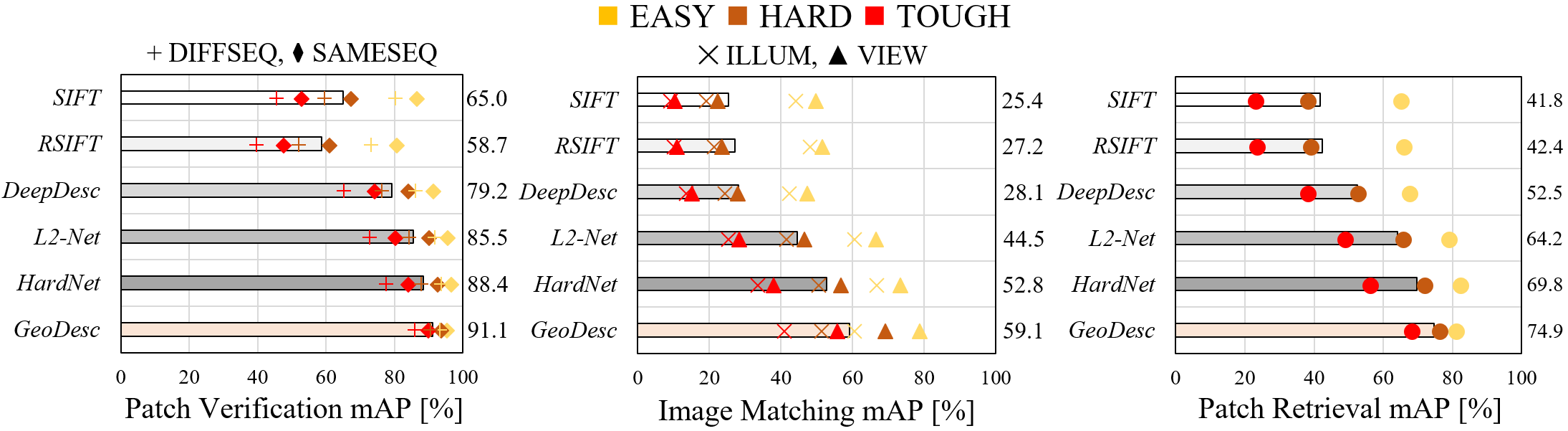}
	\caption{Left to right: Verification, matching and retrieval results on HPatches dataset, split `full'. Results on different patch groups are colorized, while DIFFSEQ/SAMESEQ and ILLUM/VIEW denote the subtasks of verification and matching, respectively}
	\label{fig:hbench}
\end{figure}

As shown in Fig.~\ref{fig:hbench}, GeoDesc surpasses all the other descriptors on all three tasks by a large margin. In particular, the performance on TOUGH patch group is significantly improved, which indicates the superior invariance to large image changes of GeoDesc. Interestingly, comparing GeoDesc with HardNet, we observe some performance drop on EASY groups especially for illumination changes, which can be ascribed to the data bias for SfM data. Though applying randomness such as illumination during the training, we cannot fully mitigate this limitation which asks more diverse real data in descriptor learning.  

In addition, we evaluate different configurations of GeoDesc on HPatches as shown in Tab.~\ref{tab:hpatches_ablation} to demonstrate the effect of each part of our method. 

\begin{itemize}[leftmargin=*, noitemsep, topsep=0pt]
	\item \emph{Config.~1}: the HardNet framework as the baseline. 
	
	\item \emph{Config.~2}: trained with the SfM data in Sec.~\ref{sec:data_praperation}. Compared with \emph{Config.~1}, it is shown that crowd-sourced training data essentially improves the generalization ability.  Meanwhile, on the other hand, \emph{Config.~2} can be regarded as an extension of LIFT~\cite{Yi:2016te} with more advanced loss formulation.
	\item \emph{Config.~3}: equipped with the proposed batch construction in Sec.~\ref{sec:batch_construction}. As discussed in Sec.~\ref{sec:loss}, the ``hardest-in-batch" strategy in HardNet is inapplicable to hard batch data and thus leads to performance drop compared with \emph{Config.~2}. In practice, we need to adjust the margin value from $1.0$ in HardNet to $0.6$, otherwise the training will not even converge. Though trainable, the smaller margin value harms the final performance. 
	\item \emph{Config.~4}: equipped with the modified structured loss in Sec.~\ref{sec:loss}. Notable performance improvements are achieved over \emph{Config.~2} due to the collaborative use of proposed methods, showing the effectiveness of simulating pair-wise matching and sampling hard data. Besides, replacing the \emph{distance margin} with \emph{distance ratio} can improve the training efficiency, as shown in Fig.~\ref{fig:fix_vs_ratio}. 
	\item \emph{Config.~5}: equipped with the geometric loss in Sec.~\ref{sec:loss}. Further improvements are obtained over \emph{Config.~4} as $E_2$ constrains the solution space and enhances the training efficiency. 
 
 \end{itemize}
 
 To sum up, the ``hardest-in-batch" strategy is beneficial when no other sampling is applied and most in-batch samples do not contribute to the loss. However, with harder batch data effectively constructed, it is advantageous to replace the ``hardest-in-batch" and further boost the descriptor performance.

 \begin{table}[ht]

 	\begin{minipage}[!ht]{0.66\linewidth}
 		
 		\centering
	\caption{Evaluation of different configurations of GeoDesc on HPatches. Modules are enabled if marked with ``Y" otherwise with ``-". \emph{SfM Data} denotes the training with our SfM data, \emph{Batch Construct.} denotes the equipment of  proposed batch construction, while $E_1$ and $E_2$ denote the use of proposed structured loss and geometric loss, respectively. The last configuration (\emph{Config. 5}) is our best model with GeoDesc}
\label{tab:hpatches_ablation}
\resizebox{\textwidth}{!}{  
	\begin{tabular}{cC{1.5cm}C{2.5cm}C{0.4cm}C{0.4cm}|C{1.7cm}C{1.4cm}C{1.3cm}}
		\Xhline{1pt}
		&\multicolumn{4}{c}{\textbf{GeoDesc Configuration}} & \multicolumn{3}{c}{\textbf{HPathces Benchmark Tasks}} \\ \Xhline{0.7pt}
		\emph{No.} &\emph{SfM Data} & \emph{Batch Construct.} & \emph{$E_1$}& \emph{$E_2$} & \emph{Verification}& \emph{Matching} & \emph{Retrieval} \\ \Xhline{0.7pt} 
		1&- & - & - & - & 88.4 & 52.8 & 69.8 \\
		2&Y & - & - & - & 90.1 & 57.0 & 73.2 \\
		3&Y & Y & - & - & 89.9 & 50.2 & 70.4 \\
		4&Y & Y & Y & - & 90.9 & 58.5 & 74.5 \\
		5&Y & Y & Y & Y & \textbf{91.1} & \textbf{59.1} & \textbf{74.9} \\
		\Xhline{1pt}
	\end{tabular}
}
 	\end{minipage}\hfill
 	 	\begin{minipage}[!ht]{0.32\linewidth}
 		\centering
 		\includegraphics[width=35mm]{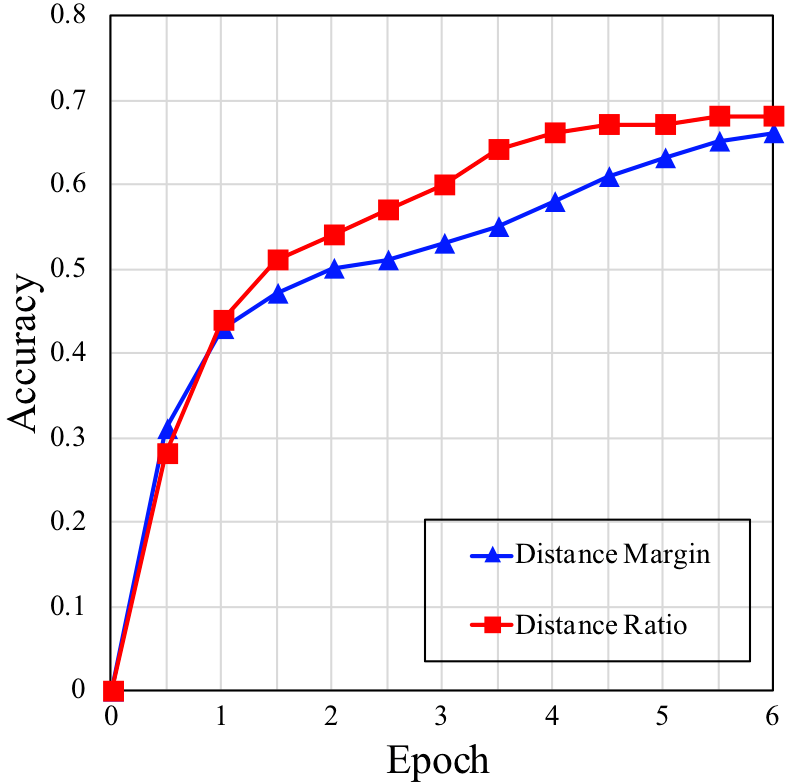}
 		\captionof{figure}{Effect of taking \emph{distance ratio} in loss computation. The metric is the validation accuracy of patch triplets with a margin of $0.5$ by cosine similarity.}
 		\label{fig:fix_vs_ratio}
 	\end{minipage}
 \end{table}
 
 \iffalse
\begin{table}[h]
	\centering
	\caption{Evaluation of different configurations of GeoDesc on HPatches benchmark. Modules are enabled if marked with ``Y" otherwise with ``-". \emph{SfM Data} denotes the training with our SfM data, \emph{Batch Construct.} denotes the equipment of  proposed batch construction, while $E_1$ and $E_2$ denote the use of proposed structured loss and geometric loss, respectively. The last configuration (\emph{Config. 5}) is our best model with GeoDesc}
	\label{tab:hpatches_ablation}
	\resizebox{\textwidth}{!}{  
		\begin{tabular}{cC{2.1cm}C{2.5cm}C{1cm}C{1cm}|C{2.cm}C{2.cm}C{2.cm}}
			\Xhline{1pt}
			&\multicolumn{4}{c}{\textbf{GeoDesc Configuration}} & \multicolumn{3}{c}{\textbf{HPathces Benchmark Task}} \\ \Xhline{0.7pt}
			\emph{No.} &\emph{SfM Data} & \emph{Batch Construct.} & \emph{$E_1$}& \emph{$E_2$} & \emph{Verification}& \emph{Matching} & \emph{Retrieval} \\ \Xhline{0.7pt} 
			1&- & - & - & - & 88.4 & 50.1 & 69.8 \\
			2&Y & - & - & - & 90.1 & 57.0 & 73.2 \\
			3&Y & Y & - & - & 89.9 & 50.2 & 70.4 \\
			4&Y & Y & Y & - & 90.9 & 58.5 & 74.5 \\
			5&Y & Y & Y & Y & \textbf{91.1} & \textbf{59.1} & \textbf{74.9} \\
			\Xhline{1pt}
		\end{tabular}
	}
\end{table}
\fi
\vspace{-10mm} 
\subsection{Heinly benchmark}
\label{sec:heinly}
Different from HPatches which experiments on image patches, the benchmark by Heinly \emph{et al.}~\cite{Heinly:2012dd} evaluates pair-wise image matching regarding different types of photometric or geometric changes, targeting to provide practical insights for strengths and weaknesses of descriptors. We use two standard metrics as in~\cite{Heinly:2012dd} to quantify the matching quality. First, the \emph{Matching Score = \#Inlier Matches / \#Features}. Second, the \emph{Recall = \#Inlier Matches / \#True Matches}. Four descriptor are selected to compare: SIFT, the baseline hand-crafted descriptor; DSP-SIFT, the best hand-crafted descriptor even superior to the previous learning-based as evaluated in~\cite{schonberger2017comparative}; L2-Net and HardNet, the recent advanced learned descriptors. For fairness comparison, no ratio test and only cross check (mutual test) is applied for all descriptors.

\begin{table}[h]
	\centering
	\caption{Evaluation results on pair-wise image matching on benchmark by Heinly \emph{et al.}~\cite{Heinly:2012dd} with respect to different types of image changes}
	\label{tab:heinly}
     \resizebox{\textwidth}{!}{
		\begin{tabular}{lccccc|ccccc}
			\Xhline{1pt}
			& \multicolumn{5}{c}{\emph{Matching Score in \%}} & \multicolumn{5}{c}{\emph{Recall in \%}}\\ \Xhline{0.7pt}
			& \textbf{SIFT} & \textbf{DSP-SIFT} & \textbf{L2-Net} & \textbf{HardNet} & \textbf{GeoDesc} & \textbf{SIFT} & \textbf{DSP-SIFT} & \textbf{L2-Net} & \textbf{HardNet} & \textbf{GeoDesc} \\ \Xhline{0.7pt}
\emph{JPEG} & 31.9 & \textbf{\color{red}35.1} & 25.7 & 27.0 & 34.7 & 60.7 & \textbf{\color{red}66.9} & 49.0 & 51.5 & 66.1 \\
\emph{Blur} & 12.4 & 14.3 & 9.1 & 11.3 & \textbf{\color{red}14.4} & 41.0 & 47.3 & 30.1 & 37.4 & \textbf{\color{red}47.7} \\
\emph{Exposure} & 32.9 & 34.8 & 33.9 & 34.9 & \textbf{\color{red}36.3} & 78.2 & 82.6 & 80.4 & 82.8 & \textbf{\color{red}86.4} \\
\emph{Day-Night} & 5.6 & 5.7 & 6.8 & 7.4 & \textbf{\color{red}7.5} & 29.2 & 29.7 & 35.6 & 38.9 & \textbf{\color{red}39.6}\\
\emph{Scale} & 35.8 & 34.7 & 32.6 & 34.8 & \textbf{\color{red}37.8} & 81.2 & 78.8 & 73.6 & 79.0 & \textbf{\color{red}85.8}\\
\emph{Rotation} & 56.3 & 49.1 & 55.9 & 57.4 & \textbf{\color{red}59.8} & 82.4 & 71.8 & 81.9 & 84.0 & \textbf{\color{red}87.6} \\
\emph{Scale-rotation} & 12.6 & 12.0 & 10.7 & 12.1 & \textbf{\color{red}14.3} & 29.6 & 28.1 & 25.0 & 28.5 & \textbf{\color{red}33.7}\\
\emph{Planar} & 23.8 & 24.8 & 25.6 & 27.4 & \textbf{\color{red}29.1} & 48.2 & 50.4 & 51.9 & 55.6 & \textbf{\color{red}59.1} \\	
\Xhline{1pt}
		\end{tabular}
		}
\end{table}

Evaluation results are shown in Tab.~\ref{tab:heinly}. Compared with DSP-SIFT, GeoDesc performs comparably regarding image quality changes (compression, blur), while notably better for illumination and geometric changes (rotation, scale, viewpoint). On the other hand, GeoDesc delivers significant improvements on L2-Net and HardNet and particularly narrows the gap in terms of photometric changes, which makes GeoDesc applicable to different scenarios in real applications.

\subsection{ETH local features benchmark}

\label{sec:eth}
The ETH local features benchmark~\cite{schonberger2017comparative} focuses on image-based 3D reconstruction tasks. We compare GeoDesc with SIFT, DSP-SIFT and L2-Net, and follow the same protocols in~\cite{schonberger2017comparative} to quantify the SfM quality, including the number of registered images (\emph{\# Registered}), reconstructed sparse points (\emph{\# Sparse Points}), image observations (\emph{\# Observations}), mean track length (\emph{Track Length}) and mean reprojection error (\emph{Reproj. Error}). For fairness comparison, we apply no distance ratio test for descriptor matching and extract features at the same keypoints as in~\cite{schonberger2017comparative}. 

As observed in Tab.~\ref{tab:eth_benchmark_sfm}, first, GeoDesc performs best on \emph{\# Registered}, which is generally considered as the most important SfM metric that directly quantifies the reconstruction completeness. Second, GeoDesc achieves best results on \emph{\# Sparse Points} and \emph{\# Observations}, which indicates the superior matching quality in the early step of SfM. However, GeoDesc fails to get best statistics about \emph{Track Length} and \emph{Reproj. Error} as GeoDesc computes the two metrics on significantly larger \emph{\# Sparse Points}. In terms of datasets whose scale is small and have similar track number (\emph{Fountain}, \emph{Herzjesu}), GeoDesc gives the longest \emph{Track Length}.

To sum up, GeoDesc surpasses both the previous best-performing DSP-SIFT and recent advanced L2-Net by a notable margin. In addition, it is noted that L2-Net also shows consistent improvements over DSP-SIFT, which demonstrates the power of taking structured loss for learned descriptors.

\begin{table*}[h]
	\centering
	\caption{Evaluation results on ETH local features benchmark~\cite{schonberger2017comparative} for SfM tasks}
	\label{tab:eth_benchmark_sfm}
	\resizebox{\textwidth}{!}{  
		\begin{tabular}{cccccccccc}
			\Xhline{1pt}
			& & \textbf{\# Images} & \textbf{\# Registered} & \textbf{\# Sparse Points} & \textbf{\# Observations} & \textbf{Track Length} & \textbf{Reproj. Error} \\ \Xhline{0.7pt}
\textbf{Fountain} & \emph{SIFT} & 11 & 11 & 10,004 & 44K & 4.49 & \textbf{\color{red}0.30px} \\
& \emph{DSP-SIFT} & & 11 & 14,785 & 71K & 4.80 & 0.41px \\
& \emph{L2-Net} & & 11 & 16,119 & 78K & 4.86 & 0.43px \\
 & \emph{GeoDesc} &  & 11 & \textbf{\color{red}16,687} & \textbf{\color{red}83K} & \textbf{\color{red}4.99} & 0.46px \\ \Xhline{0.7pt}
\textbf{Herzjesu} & \emph{SIFT} & 8 & 8 & 4,916 & 19K & 4.00 & \textbf{\color{red}0.32px} \\
& \emph{DSP-SIFT} & & 8 & 7,760 & 32K & 4.19 & 0.45px \\
& \emph{L2-Net} & & 8 & 8,473 & 36K & 4.27 & 0.47px \\
 & \emph{GeoDesc} &  & 8 & \textbf{\color{red}8,720} & \textbf{\color{red}38K} & \textbf{\color{red}4.34} & 0.55px \\ \Xhline{0.7pt}
\textbf{South Building} & \emph{SIFT} & 128 & 128 & 62,780 & 353K & 5.64 & \textbf{\color{red}0.42px} \\
& \emph{DSP-SIFT} & & 128 & 110,394 & 664K & \textbf{\color{red}6.02} & 0.57px \\
& \emph{L2-Net} & & 128 & 155,780 & 798K & 5.13  & 0.58px \\
& \emph{GeoDesc} & & 128 & \textbf{\color{red}170,306} & \textbf{\color{red}887K} & 5.21 & 0.64px \\ 
\Xhline{0.7pt}
\textbf{Madrid Metropolis} & \emph{SIFT} & 1,344 & 440 & 62,729 & 416K & \textbf{\color{red}6.64} & \textbf{\color{red}0.53px} \\
& \emph{DSP-SIFT} & & 476 & 107,028 & 681K & 6.36 & 0.64px \\
& \emph{L2-Net} & & 692 & 254,142 & 1,067K & 4.20 & 0.69px \\
& \emph{GeoDesc} & & \textbf{\color{red}809} & \textbf{\color{red}306,976} & \textbf{\color{red}1,200K} & 3.91 & 0.66px \\
\Xhline{0.7pt}
\textbf{Gendarmenmarkt} & \emph{SIFT} & 1,463 & 950 & 169,900 & 1,010K & \textbf{\color{red}5.95} & \textbf{\color{red}0.64px} \\
& \emph{DSP-SIFT} & & 975 & 321,846 & 1,732K & 5.38 & 0.74px \\
& \emph{L2-Net} & & 1,168 & 667,392 & 2,611K & 3.91 & 0.73px \\
& \emph{GeoDesc} & & \textbf{\color{red}1,208} & \textbf{\color{red}779,814} & \textbf{\color{red}2,903K} & 3.72 & 0.74px \\
\Xhline{0.7pt}
\textbf{Tower of London} & \emph{SIFT} & 1,576 & 702 & 142,746 & 963K & 6.75 & \textbf{\color{red}0.53px} \\
& \emph{DSP-SIFT} & & 755 & 236,598 & 1,761K & \textbf{\color{red}7.44} & 0.64px \\
& \emph{L2-Net} & & 1,049 & 558,673 & 2,617K & 4.68 & 0.67px \\
& \emph{GeoDesc} & & \textbf{\color{red}1,081} & \textbf{\color{red}622,076} & \textbf{\color{red}2,852K} & 4.58 & 0.69px \\
\Xhline{0.7pt}
\textbf{Alamo} & \emph{SIFT} & 2,915 & 743 & 120,713 & 1,384K & 11.47 & \textbf{\color{red}0.54px} \\
& \emph{DSP-SIFT} & & 754 & 144,341 & 1,815K & \textbf{\color{red}12.58} & 0.66px \\
& \emph{L2-Net} & & 882 & 318,787 & 2,932K & 9.17 & 0.76px \\
& \emph{GeoDesc} & & \textbf{\color{red}893} & \textbf{\color{red}353,329} & \textbf{\color{red}3,159K} & 8.94 & 0.84px \\
\Xhline{0.7pt}
\textbf{Roman Forum} & \emph{SIFT} & 2,364 & 1,407 & 242,192 & 1,805K & 7.45 & \textbf{\color{red}0.61px} \\
& \emph{DSP-SIFT} & & \textbf{\color{red}1,583} & 372,573 & 2,879K & \textbf{\color{red}7.73} & 0.71px \\
& \emph{L2-Net} & & 1,537 & 708,794 & 4,530K & 6.39 & 0.69px \\
& \emph{GeoDesc} & & 1,566 & \textbf{\color{red}770,363} & \textbf{\color{red}5,051K} & 6.56 & 0.73px \\
\Xhline{0.7pt}
\textbf{Cornell} & \emph{SIFT} & 6,514 & 4,999 & 1,010,544 & 6,317K & \textbf{\color{red}6.25} & \textbf{\color{red}0.53px} \\
& \emph{DSP-SIFT} & & 4,946 & 1,177,916 & 7,233K & 6.14 & 0.67px & \\
& \emph{L2-Net} & & 5,557 & 2,706,215 & 15,710K & 5.81 & 0.72px & \\
& \emph{GeoDesc} & & \textbf{\color{red}5,823} & \textbf{\color{red}3,076,476} & \textbf{\color{red}17,550K} & 5.70 & 0.96px & \\
\Xhline{0.7pt}
		\end{tabular}
	}
\end{table*}

\subsection{Challenging 3D reconstructions}

\label{sec:sfm}

To further demonstrate the effect of the proposed descriptor in a context of 3D reconstruction, we showcase selective image sets whose reconstructions fail or are in low quality with a typical SIFT-based 3D reconstruction pipeline but get significantly improved by integrating GeoDesc.

\begin{figure}[H]
	\centering 
	\includegraphics[width=\textwidth]{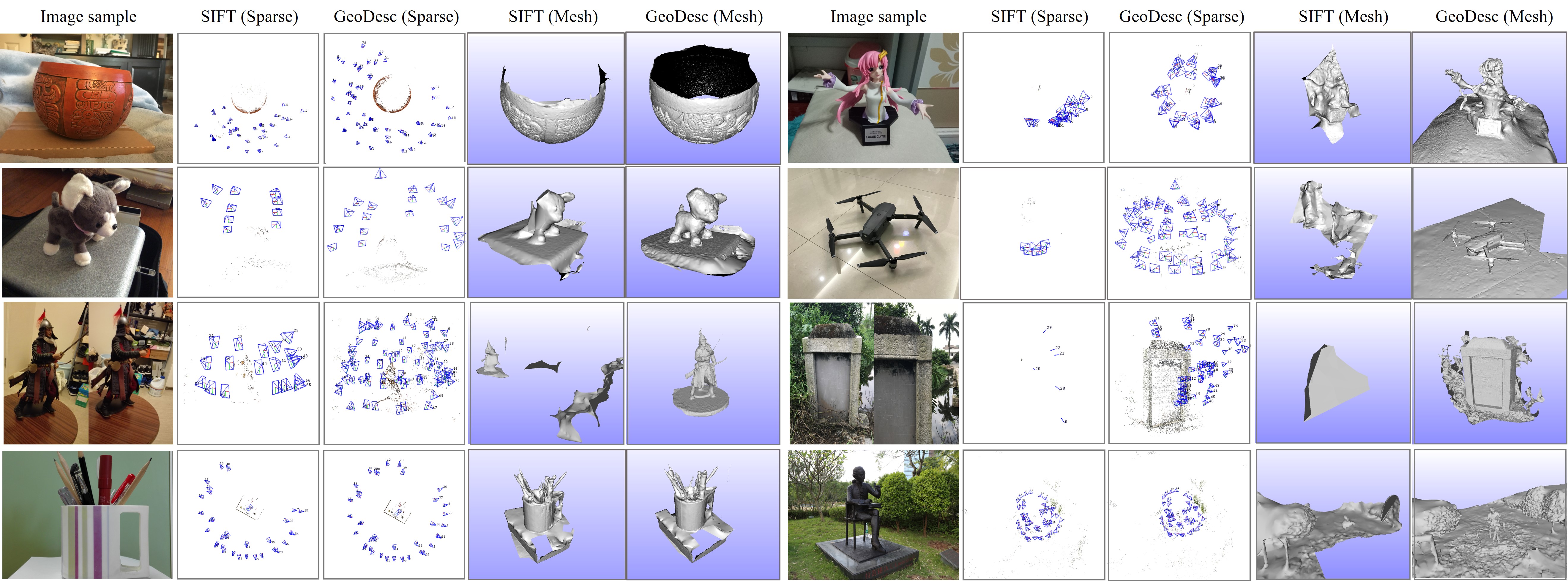}
	\caption{Testing cases of challenging image sets, where a traditional SIFT-based reconstruction
		pipeline fails to apply but GeoDesc delivers significant improvement.}
	\label{fig:challenging_sfm}
\end{figure}

From examples shown in Fig.~\ref{fig:challenging_sfm}, it is clear to see the benefit of deploying GeoDesc in a reconstruction pipeline. First, by robust matching resistant to photometric and geometric changes, a complete sparse reconstruction registered with more cameras can be obtained. Second, due to more accurate camera pose estimation, the final fined mesh reconstruction is then derived.

\section{Practical Guidelines}
In this section, we discuss several practical guidelines to complement the performance evaluation and provide insights towards real applications. Following experiments are conducted with $231$ extra high-resolution image pairs, whose keypoints are downsampled to $\sim$10k per image. We use a single NVIDIA GTX 1080 GPU with TensorFlow~\cite{abadi2016tensorflow}, and forward each batch with 256 patches. 

\subsection{Ratio criterion}
\label{sec:ratio}
The ratio criterion~\cite{Lowe:2004kp} compares the distance between the first and the second nearest neighbor, and establishes a match if the former is smaller than the latter to some ratio. For SfM tasks, the ratio criterion improves overall matching quality, RANSAC efficiency, and seeds robust initialization. Despite those benefits, the ratio criterion has received little attention, or even been considered inapplicable to learned descriptors in previous studies~\cite{schonberger2017comparative}. Here, we propose a  general method to determine the ratio that well cooperates with existing SfM pipelines.

The general idea is simple: the new ratio should function similarly as SIFT's, as most SfM pipelines are parameterized for SIFT. To quantify the effect of the ratio criterion, we use the metric \emph{Precision = \#Inlier Matches / \#Putative matches}, and determine the ratio that achieves similar \emph{Precision} as SIFT's. As an example in Fig.~\ref{fig:ratio}, we compute the \emph{Precision} of SIFT and GeoDesc on our experimental dataset, and find the best ratio for GeoDesc is $0.89$ at which it gives similar \emph{Precision} ($0.70$) as SIFT ($0.69$). This ratio is applied to experiments in Sec.~\ref{sec:sfm} and shows robust results and compatibility in the practical SfM pipeline.
   
\subsection{Compactness study}
\label{sec:compactness}
A compact feature representation generally indicates better performance with respective to discriminativeness and scalability. To quantify the compactness, we reply on the intermediate result in Principal Component Analysis (PCA). First, we compute the explained variance $v_i$ which is stored in increasing order for each feature dimension indexed by $i$. Then we estimate the compact dimensionality (denoted as Compact-Dim) by finding the minimal $k$ that satisfies $\sum^k_i{v_i} / \sum^D_i{v_i} >= t$, where $t$ is a given threshold and $D$ is the original feature dimensionality. In this experiment, we set $t = 0.9$, so that the Compact-Dim can be interpreted as the minimal dimensionality required to convey more than $90\%$ information of the original feature. Obviously, larger Compact-Dim indicates less redundancy, namely greater compactness.

As a result, the Compact-Dim estimated on 4 millions feature vectors for SIFT, DSP-SIFT, L2-Net and GeoDesc is 56, 63, 75 and 100, respectively. The ranking of Compact-Dim effectively responds to previous performance evaluations, where descriptors with larger Compact-Dim yield better results.

\subsection{Scalability study}
\noindent\textbf{Computational cost.} As evaluated in~\cite{Balntas:2017vx,schonberger2017comparative}, the efficiency of learned descriptors is on par with traditional descriptors such as CPU-based SIFT. Here, we further compare with GPU-based SIFT~\cite{wu2007siftgpu} to provide insights about practicability. We evaluate the running time in three steps. First, keypoint detection and canonical orientation estimation by SIFT-GPU. Next, patches cropping by Equ.~\ref{eq:sampling}. Finally, feature inference of image patches. The computational cost and memory demand are shown in Tab.~\ref{tab:efficiency}, indicating that with GPU support, not surprisingly, SIFT ($0.20s$) is still faster than the learned descriptor ($0.31s$), with a narrow gap due to the parallel implementation. For applications heavily relying on  matching quality (e.g., 3D reconstruction), the proposed descriptor achieves a good trade-off to replace SIFT.

\smallskip\noindent\textbf{Quantization.} To conserve disk space, I/O and memory, we linearly map feature vectors of GeoDesc from $[-1, 1]$ to $[0, 255]$ and round each element to unsigned-char value. The quantization does not affect the performance as evaluated on HPatches benchmark.

\begin{table}[ht]
\begin{minipage}[!ht]{0.4\linewidth}
\centering
\includegraphics[width=33mm]{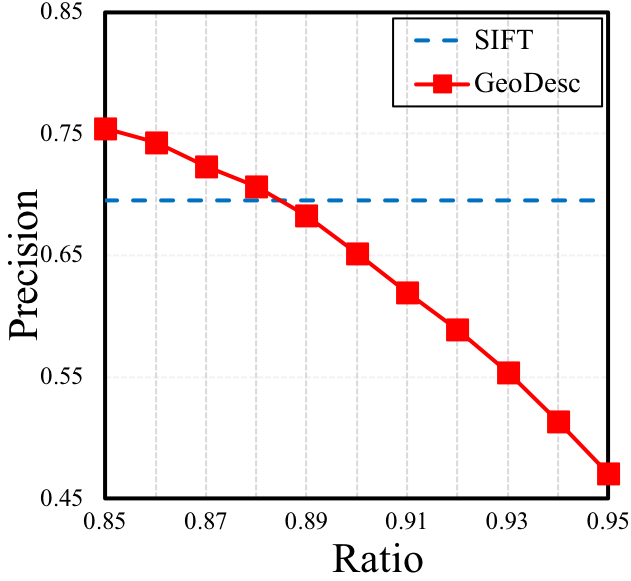}
\captionof{figure}{Determine the ratio criterion of GeoDesc so that it has the same \emph{Precision} as SIFT (at $0.89$)}
\label{fig:ratio}
\end{minipage}\hfill
\begin{minipage}[!ht]{0.6\linewidth}

\centering
\caption{Computational cost and memory demand of feature extraction of GeoDesc in three steps: SIFT-GPU extraction, patch cropping and feature inference. The total time cost is evaluated with three steps implemented in a parallel fashion}
\resizebox{0.70\textwidth}{!}{
		\begin{tabular}{ccccc}
			\Xhline{1pt}
			& \textbf{SIFT} & \textbf{Crop.} & \textbf{Infer.} & \textbf{Total} \\ \Xhline{0.7pt}
			\emph{Device} & GPU & CPU & GPU & - \\
			\emph{Memory (GB)} & 3.3 & 2.7 & 0.3 & - \\			\emph{Time (s)} & 0.20 & 0.28 & 0.31 & 0.31 \\

			\Xhline{1pt}
		\end{tabular}
	}
    
    \label{tab:efficiency}
\end{minipage}

\end{table}

\section{Conclusions}
In contrast to prior work, we have addressed the advantages of integrating geometry constraints for descriptor learning, which benefits the learning process in terms of ground truth data generation, data sampling and loss computation. Also, we have discussed several guidelines, in particular, the \emph{ratio criterion}, towards practical portability. Finally, we have demonstrated the superior performance and generalization ability of the proposed descriptor, GeoDesc, on three benchmark datasets in different scenarios, We have further shown the significant improvement of GeoDesc on challenging reconstructions, and the good trade-off between efficiency and accuracy to deploy GeoDesc in real applications.

\clearpage

\bibliographystyle{splncs}
\bibliography{egbib}
\end{document}